\documentclass[journal,twoside,web]{ieeecolor}

\usepackage{etoolbox}

\makeatletter
\@ifundefined{color@begingroup}%
{\let\color@begingroup\relax
\let\color@endgroup\relax}{}%
\def\fix@ieeecolor@hbox#1{%
\hbox{\color@begingroup#1\color@endgroup}}
\patchcmd\@makecaption{\hbox}{\fix@ieeecolor@hbox}{}{\FAILED}
\patchcmd\@makecaption{\hbox}{\fix@ieeecolor@hbox}{}{\FAILED}

\usepackage{generic}
\usepackage{cite}

\usepackage{amsmath,amssymb,amsfonts}
\usepackage{graphicx}
\usepackage{textcomp}
\usepackage{siunitx}
\usepackage{algorithm}
\usepackage{CJK}
\usepackage{algorithmicx}
\usepackage{algpseudocode}
\usepackage{multirow}
\usepackage{threeparttable}
\usepackage{lscape}
\usepackage{booktabs}

\usepackage{tikz}
\usepackage[implicit=false]{hyperref}
\hypersetup{hidelinks,
	colorlinks=true,
	allcolors=black,
	pdfstartview=Fit,
	breaklinks=true}
\definecolor{lime}{HTML}{A6CE39}
\DeclareRobustCommand{\orcidicon}{
\begin{tikzpicture}
\draw[lime, fill=lime] (0,0)
circle[radius=0.16]
node[white]{{\fontfamily{qag}\selectfont \tiny \.{I}D}};
\end{tikzpicture}
\hspace{-2mm}
}
\foreach \x in {A, ..., Z}{%
\expandafter\xdef\csname orcid\x\endcsname{\noexpand\href{https://orcid.org/\csname orcidauthor\x\endcsname}{\noexpand\orcidicon}}
}

\usepackage{makecell}

\def\BibTeX{{\rm B\kern-.05em{\sc i\kern-.025em b}\kern-.08em
    T\kern-.1667em\lower.7ex\hbox{E}\kern-.125emX}}
    
\begin{document}
\title{Using Legacy Polysomnography Data to Train a Radar System to Quantify Sleep in Older Adults and People living with Dementia}
\author{Maowen Yin \hspace{-1.5mm}\orcidD{}, Kiran K G Ravindran \hspace{-1.5mm}\orcidH{}, Charalambos Hadjipanayi \hspace{-1.5mm}\orcidE{},  Alan Bannon \hspace{-1.5mm}\orcidF{}, Ciro della \\ \vspace{-0.15cm} Monica \hspace{-1.5mm}\orcidI{}, Adrien Rapeaux \hspace{-1.5mm}\orcidC{},  Tor Sverre Lande \hspace{-1.5mm}\orcidB{}, \IEEEmembership{Fellow, IEEE}, Derk-Jan Dijk \hspace{-1.5mm}\orcidG{}and Timothy \\ \vspace{-0.15cm} Constandinou \hspace{-1.5mm}\orcidA{}, \IEEEmembership{Senior Member, IEEE} \vspace{-0.5cm}
\thanks{"This work is supported by the UK Dementia Research Institute [award number UK DRI-7204 and UK DRI-7206] through UK DRI Ltd, principally funded by the UK Medical Research Council, and additional funding partner the British Heart Foundation." }
\thanks{M. Yin, C. Hadjipanayi, A. Bannon, A. Rapeaux, and T. G. Constandinou are with the Department of Electrical and Electronic Engineering and UK Dementia Research Institute (Care Research \& Technology Centre), Imperial College London, SW7 2BT, United Kingdom (e-mail: maowen.yin20@imperial.ac.uk). }
\thanks{Tor Sverre Lande is with the Department of Informatics, University of Oslo, Oslo 0315, Norway. }
\thanks{Kiran K G Ravindran, Ciro della Monica, and Derk-Jan Dijk are with the School of Biosciences, University of Surrey, and UK Dementia Research Institute (Care Research \& Technology Centre), University of Surrey. }}

\maketitle

\begin{abstract}

Objective: Ultra-wideband (UWB) radar technology offers a promising solution for unobtrusive and cost-effective in-home sleep monitoring. However, the limited availability of radar sleep data poses challenges in building robust models that generalize across diverse cohorts and environments. This study proposes a novel deep transfer learning framework to enhance sleep stage classification using radar data.
Methods: An end-to-end neural network was developed to classify sleep stages based on nocturnal respiratory and motion signals. The network was trained using a combination of large-scale polysomnography (PSG) datasets and radar data. A domain adaptation approach employing adversarial learning was utilized to bridge the knowledge gap between PSG and radar signals. Validation was performed on a radar dataset of 47 older adults (mean age: 71.2 ± 6.5), including 18 participants with prodromal or mild Alzheimer’s disease.
Results: The proposed network structure achieves an accuracy of 79.5\% with a Kappa value of 0.65 when classifying wakefulness, rapid eye movement, light sleep and deep sleep.
Experimental results confirm that our deep transfer learning approach significantly enhances automatic sleep staging performance in the target domain.
Conclusion: This method effectively addresses challenges associated with data variability and limited sample size, substantially improving the reliability of automatic sleep staging models, especially in contexts where radar data is limited.
Significance: The findings underscore the viability of UWB radar as a nonintrusive, forward-looking sleep assessment tool that could significantly benefit care for older people and people with neurodegenerative disorders.

\end{abstract}


\section{Introduction}
\label{sec:introduction}

\begin{figure*}
\centering
\includegraphics[width=0.95\textwidth]{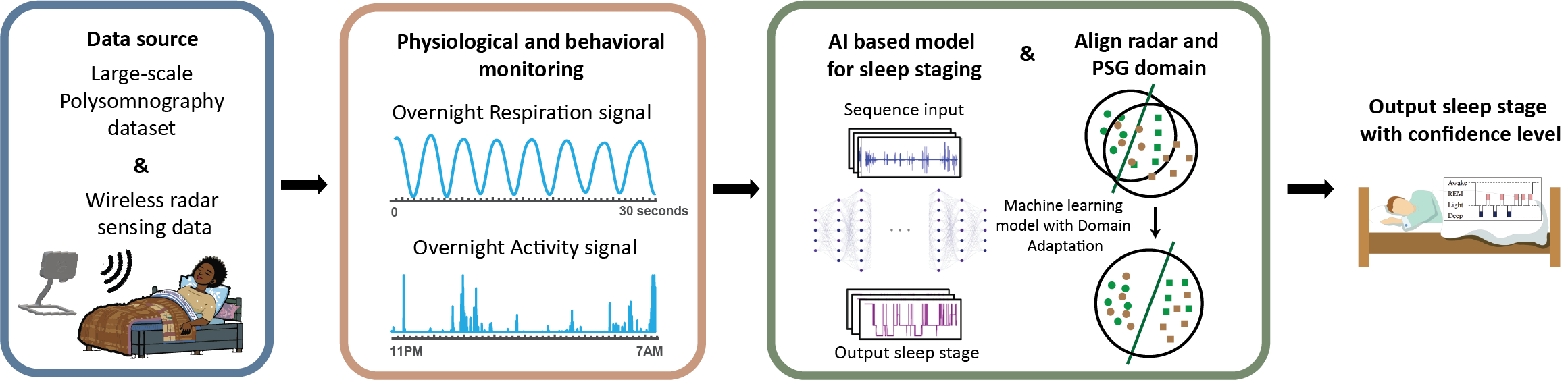}
\caption{ Overview of the proposed approach for radar-based sleep staging and transfer learning. The nighttime activity signal and respiration signal are used to predict the sleep stage and a domain adaptation model is used to align feature distributions between the source domain PSG data and the target domain radar signals. }
\label{transfer_therory}
\end{figure*}

The importance of sleep monitoring in healthcare is increasingly recognized.  Adequate sleep is fundamental to well-being, and disruptions are linked to various health anomalies \cite{gottesman2024impact}. A comprehensive, longitudinal, and objective sleep monitoring system can help users identify sleep issues early or track the progression of sleep disturbance in brain disorders such as dementia. According to guidelines from the American Academy of Sleep Medicine (AASM), sleep architecture is divided into five distinct stages: 1) wakefulness (W); 2) rapid eye movement (REM); 3) non-rapid eye movement stage 1 (N1); 4) non-REM stage 2 (N2); 5) non-REM stage 3 (N3) \cite{silber2007visual}. In most evaluation studies of novel sleep monitoring devices N1 and N2 are combined as light sleep and N3 is referred to as deep sleep.

Although polysomnography (PSG) remains the gold standard for sleep assessment, its scalability for long-term use is limited due to its obtrusive nature and high costs. Its application in longitudinal monitoring of older people with neurodegenerative disorders is particularly challenging due to increased sensor discomfort and compliance issues. This challenge has directed research towards developing unobtrusive, cost-effective sleep monitoring methods. In this context, radar technology has emerged as a viable alternative, offering the capability for remote sensing of physiological and behavioral signals. 

Radar sensing relies on the Doppler effect, which is produced by the relative displacement between the radar and the person. When the person moves, the radar captures varying phase-shifted echoes reflected from different body parts, which can be used to distinguish different types of activity or perform gait analysis \cite{10518096,bannon2021tiresias}. When the person is stationary, the radar's high sensitivity allows for the detection of chest displacements caused by breathing and heartbeats \cite{9441392}. Radar technology has shown higher accuracy compared with smartwatches or smart sleep-monitoring mattresses in sleep staging \cite{g2023three}. Moreover, the radar's contactless nature, requiring no charging or maintenance, significantly enhances user compliance, especially among older adults and those with neurodegenerative diseases.

Recent advancements in non-contact sleep monitoring have significantly developed from nighttime physiological signal monitoring to more in-depth sleep stage classification \cite{rahman2015dopplesleep,vasireddy2018k,hu2013noncontact}. Influenced by nervous system regulation, distinct sleep stages exhibit significant differences in movement, breathing, and heart rate signals \cite{malik2012respiratory, somers1993sympathetic}. For instance, deeper sleep is associated with increased parasympathetic activity, resulting in more regular heart and breathing rhythms, as well as shallower breathing. Early research in this field utilized manually extracted features from radar signals for sleep stage classification using traditional machine-learning techniques \cite{heglum2021distinguishing,de2021radar}. The latest advancements in artificial intelligence have driven the use of deep learning for sleep staging. Convolutional neural networks (CNN) were used for extracting sleep features and recurrent neural networks (RNN) were used for learning the relationship between adjacent periods \cite{toften2020validation}. This end-to-end network structure has significantly improved the accuracy of sleep stage classification. More recently, an expanded multi-center study incorporating a broader and more heterogeneous cohort further demonstrated that radar can achieve strong sleep-staging performance across diverse environments and populations, reinforcing its translational potential beyond highly controlled laboratory settings \cite{he2025radio}.

However, deep learning models require large, well-annotated datasets that reflect consistent feature distributions across training and testing sets. Most published studies on radar-based sleep monitoring have been limited to small cohorts (\textless 100 participants), predominantly healthy adults, and conducted in controlled sleep laboratories with ideal radar placements. This results in models that may not perform accurately with new subjects or under different operational conditions \cite{baumert2023automatic}. This reveals a significant gap: the absence of large-scale, diverse datasets that integrate radar-based observations with traditional PSG data.

Transfer learning is a potential solution to this challenge, aiming to leverage large datasets' prior knowledge for better model training \cite{pan2009survey}. In the context of sleep classification using radar, this allows for the option of pre-training a classifier on a labeled dataset based on physiological signals collected using PSG. However, physiological signals extracted by radar differ from those collected by PSG. To address this issue, we employed a domain adaptation model (principles shown in Fig. \ref{transfer_therory}) that aligns feature distributions between the source domain PSG data and the target domain radar signals, minimizing domain shift \cite{li2018cross}.

This study aims to improve the accuracy of sleep stage classification based on radar signals through transfer learning. We propose an end-to-end deep neural network that includes a CNN-based feature extractor, a sleep stage classifier, and an adversarial learning-based domain adaptation module, all of which can be trained jointly. 
The proposed method was subjected to a challenging validation in a small scale trial in older participants ($N$=47, mean age 71.2 ± 6.5 years), including 18 participants living with prodromal or mild Alzheimer’s disease (PLWAD). The results show that it significantly enhances the accuracy of radar-based sleep classification on small datasets and improves the robustness of the model across different subjects. 
Our contributions include:

\begin{itemize}
    \item Developing a novel transfer learning technique to enhance the performance of radar-based sleep monitoring.
    \item Introducing a discriminator to overcome domain differences between radar and PSG signals, producing domain-invariant features.
    \item Evaluating in a diverse cohort of older adults and PLWAD, alongside varying radar placements, thereby characterizing the model's robustness across different participants and sensing setups.
\end{itemize}

\section{Materials} \label{setup}

\begin{figure}
\centering
\includegraphics[width=0.45\textwidth]{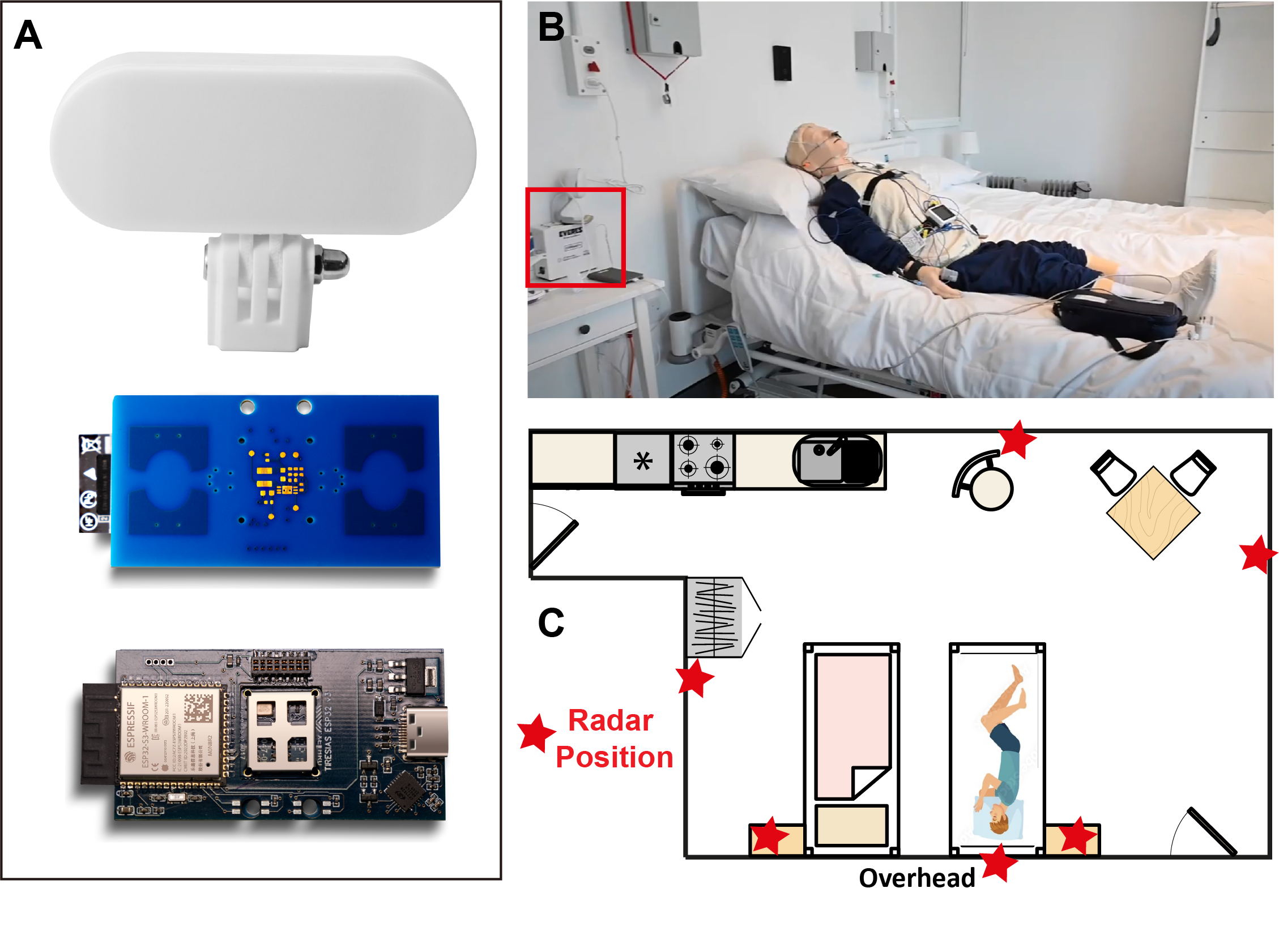}
\caption{ The setup of the experiment: Radar recording system (Fig A\&B) and floor map of the UK DRI clinical research facility at the University of Surrey (Fig C\&D). The position of the radar is marked with a red pentagram. Figure reuse from \cite{yin2025unobtrusive}.}
\label{structure}
\end{figure}

\subsection{Source Domain} \label{EVE}

In this study, the Stanford Technology Analytics and Genomics in Sleep (STAGES) database was utilized as the source domain due to its extensive data coverage \cite{zhang2018national}.
The STAGES database consists of a comprehensive multi-center study with full-night PSG recordings from 1,662 individuals, following the exclusion of incomplete datasets. The participants had a mean age of 45.8 years, with a standard deviation of 15.2 years. The dataset includes various physiological and movement measurements, such as electroencephalograms (EEGs), electromyograms (EMGs) from the chin and legs, nasal and oral airflow, chest and leg movements, body position, electrocardiograms (ECGs), and data from inertial measurement units (IMUs).

For this study, the respiratory signal used for analysis was derived from a chest respiratory band, which employed inductive plethysmography and sampled at 50 Hz. The IMU data, recorded at a frequency of 25 Hz, captured three-axis accelerometer readings from a wrist-worn device, with a resolution of 12 bits and a dynamic range of ±8 g. Due to missing data, either from interruptions during nighttime recording or incomplete recordings of the respiratory band or IMU data, 197 participants were excluded from the final analysis.


\begin{table}
\centering
\caption{Summary of demographics and sleep parameters}
\begin{threeparttable}
\begin{tabular}{llll}
\toprule

Characteristics  & STAGES & Surrey Phase I & Surrey Phase II\\ \midrule
N & 1662 & 11 & 36\\
Radar No. & -   &  1  & 3-6 \\
Sex (M/F)           & $786 / 876$       & $7 / 4$           & $21 / 15$\\
Age (years)         & $45.9 \pm 15.2$   & $71.3 \pm 5.0$    & $70.5 \pm 8.4$\\
BMI ($kg/{m}^2$)    & $27.2 \pm 6.5$    & $25.1 \pm 3.6$    & $26.8 \pm 5.6$\\
AHI (events/h)      & $13.5 \pm 19.2$   & $15.6 \pm 8.7$    & $17.5 \pm 13.7$\\
PLMI (events/h)     & $7.0 \pm 18.1$    & $22.5 \pm 34.4$   & $24.5 \pm 15.6$\\
TST (min)           & $354.1 \pm 89.5$  & $333.8 \pm 48.0$  & $321.0 \pm 72.0$\\
SE (\%)             & $76.4 \pm 15.1$   & $65.2 \pm 11.0$   & $66.6 \pm 14.3$\\
SoL (min)           & $29.3 \pm 37.7$   & $22.9 \pm 14.8$   & $16.1 \pm 23.0$\\
WASO (min)          & $91.7 \pm 72.3$   & $166.2 \pm 58.9$  & $153.2 \pm 51.6$\\
Stage N1 (\%)       & $11.7 \pm 10.2$   & $21.1 \pm 9.6$    & $15.9 \pm 7.8$\\
Stage N2 (\%)       & $62.9 \pm 24.9$   & $45.9 \pm 8.0$    & $44.0 \pm 9.3$\\
Light sleep (\%)    & $70.7 \pm 16.9$   & $66.9 \pm 13.1$   & $60.0 \pm 11.6$\\
Deep sleep (\%)     & $10.7 \pm 10.51$     & $20.1 \pm 11.5$   & $25.9 \pm 10.7$\\
REM (\%)            & $18.6 \pm 7.2$    & $12.9 \pm 5.3$    & $14.3 \pm 6.0$\\
\bottomrule

\end{tabular}
\label{parameter}

\begin{tablenotes}
\small
\item Note: S.D.: Standard deviation; BMI: Body mass index; AHI: Apnea-hypopnea index; PLMI: Periodic limb movement index; TST: Total sleep time; WASO: Wake after sleep onset; SE: Sleep efficiency; SOL: Sleep Onset Latency; Light sleep, summation of stages N1 and N2; Deep sleep, N3 stage.
\end{tablenotes}
\end{threeparttable}
\label{Anthropometric2}
\end{table}

\subsection{Target Domains}

This target domain consisted of a small radar sleep measurement dataset. Our work uses the system-on-chip (SoC) radar transceiver, the {\textit{XeThru X4}}, manufactured by {\textit{Novelda AS, Oslo, Norway}}. This sensor is CE-certified and the emission energy is below the Federal Communications Commission (FCC) compliance, which is essential for indoor and healthcare applications. The radar sensor was placed on a bedside table, facing the participants, and was used for continuous monitoring throughout the night (from lights off to lights on), as illustrated in Fig. \ref{structure}.

The radar sleep study was conducted at the UK DRI Clinical Research Facility at the University of Surrey to collect radar data and validate the proposed methods. This facility is designed to study participants in an environment where they can engage in activities during the daytime and sleep at night while being monitored by clinical and research staff. Participants spent an entire day in the laboratory, including an overnight stay where their usual sleep and wake times were adhered to. The study was reviewed and approved by the University of Surrey Ethics Committee (approval ref: UEC 2019 065 FHMS) and the London City \& East Research Ethics Committee (approval ref: 22/LO/0694). It was carried out in accordance with the Declaration of Helsinki, the Principles of Good Clinical Practice, and relevant University of Surrey guidelines and regulations. Trained staff performed the data collection, and all participants provided written informed consent before data collection. The detailed protocol can be found in \cite{della2023protocol}, with anthropometric and sleep parameters summarized in Table \ref{Anthropometric2}.

The study was divided into two phases. Clinical PSG was used as the gold standard for sleep measurements in both phases, which were recorded simultaneously with radar data collection.

\subsubsection{Participant Overview}

\textbf{Phase I:} 
Eleven cognitively healthy elderly participants were included in the first phase. Each participant spent the night in a private room during the data collection.
\textbf{Phase II:} 
The second phase involved 36 participants, including 18 individuals with prodromal or mild Alzheimer’s disease, 11 caregivers, and 7 healthy controls. When caregivers participated alongside the individuals with Alzheimer’s, they both shared the room, sleeping on adjacent beds.

\subsubsection{Radar Placement}
\textbf{Phase I:} 
A single radar device was placed on the bedside table for data collection. 
\textbf{Phase II:}  
Multiple radar devices with different positions in the room were used to capture a wider array of data points. These devices included one placed on the bedside table, one overhead, and three on the side wall, as illustrated in Fig. 2.

\section{Method}

\subsection{PSG signal processing}

The Surrey PSG data was used solely for synchronization with radar data and manual sleep stage scoring, serving as a reference, and was not used for training. 

In both the STAGES and Surrey cohorts, overnight polysomnography was segmented into 30-s epochs and scored by registered sleep technologists according to the American Academy of Sleep Medicine (AASM) criteria. The five standard stages (W, N1, N2, N3, and REM) were consolidated into four classes for analysis: wake (W), rapid eye movement (REM), light sleep (LS, N1+N2), and deep sleep (DS, N3).

Given that radar cannot directly observe brain activity, it infers sleep stages from changes in motion and physiological signals. Thus, the PSG channels used here only included a respiratory signal from a chest belt and an activity signal (ACT) from a wrist-worn IMU. 

The respiratory signal went through a band-pass filter with a cutoff frequency of 6 to 36 breaths per minute \cite{takayama2019aging}. For each subject, respiratory signals from the entire night were globally normalized to have zero mean and unit standard deviation. The respiratory signal was then downsampled to 10 Hz for data size reduction and stored for input to the neural network. 

The activity signal was derived from the magnitude of the acceleration vector calculated as:

\begin{equation}
\begin{aligned}
r=\sqrt{\left(a c c^x\right)^2+\left(a c c^y\right)^2+\left(a c c^z\right)^2}
\end{aligned}
\end{equation}

The raw IMU signal was then passed through a band-pass filter with a range between 0.25 Hz to 2.5 Hz to remove low-frequency components and high-frequency noise. The filtered activity signal was downsampled to 10 Hz by an overlapped integration window. Each data point is an integration over $1s$ interval $n$, the $\mathrm{ACT}_t$ is

\begin{equation}
\begin{aligned}
\mathrm{ACT}_t=\frac{1}{M} \sum_{i=t}^{t + M}\left|r_i-\left\langle r_i\right\rangle\right|
\end{aligned}
\end{equation}

where $M$ is the number of points in the 1s interval and $\left\langle r_i\right\rangle$ is the corresponding average. 

\subsection{Radar signal processing}

Similar to PSG signals, radar signals undergo preprocessing to extract two data channels: the activity signal and the respiratory signal, which reflect motion and physiological changes, respectively. These signals are stored as 1D signals at a sampling rate of 10 Hz and synchronized with the Surrey PSG data using time stamps. For each subject, signals collected throughout the night are also globally normalized per subject.

\subsubsection{Signal pre-processing}

UWB radar operates by transmitting a series of electromagnetic pulses $p(\tau)$ toward an object, and the modulated backscattered pulses $r(t, \tau)$ are then measured to estimate movement between the target and the transmission source. 

\begin{equation}
r(t, \tau)=\sum_i A_i p\left(\tau-\tau_i\right)+A_p \cdot p\left(\tau-\tau_d(t)\right)
\end{equation}

where $r(t, \tau)$ indicates the received signal with a real-time factor t, which is generally called slow-time, and the observed range factor $\tau$, which is called fast-time. $p(\tau)$ is the normalized received pulse. where $A_p$ and $\tau_d(t)$ are the direct reflection amplitude and propagation time due to cardiopulmonary activity. $A_i$ and $\tau_i$ refer to the reflection amplitude and time delay from the static background targets, called clutter, which are suppressed using an adaptive filter. 

\begin{figure}
\centering
\includegraphics[width=0.5\textwidth]{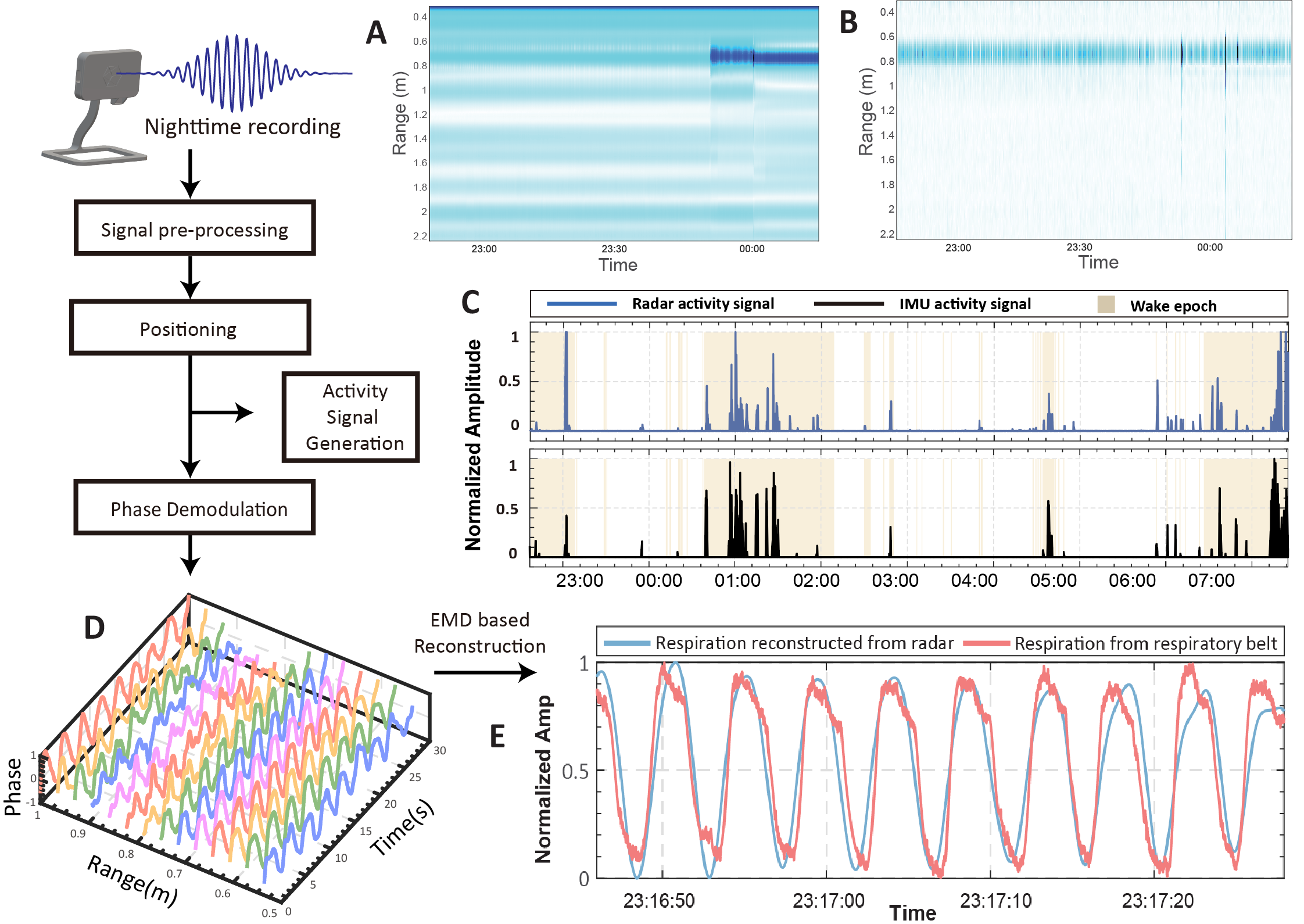}
\caption{  Signal pre-processing and respiration reconstruction pipeline. A: initial range profile of the raw radar signal, and B: range profile after clutter removal. C: Radar-derived and IMU-derived activity signal. D: Phase-demodulated range–time map within the participant’s range of interest. E: 30-second comparison between radar $R(t)$ and belt data. }
\label{pre-processing}
\end{figure}

Fig.~\ref{pre-processing}A illustrates a two-hour segment of raw range--time radar data from one participant. After clutter removal (Fig.~\ref{pre-processing}B), most static reflections, including those from stationary body parts, are attenuated, leaving predominantly motion-related components.

\subsubsection{Position estimation and range filtering} \label{Position}

Impulse radar can determine the distance to the participants and filter out further noise, clutter, and multipath sources by selecting a suitable observation range. Given that the respiration signal frequency mostly falls within the  6 to 36 breaths per minute frequency range \cite{takayama2019aging}, the position of the participant can be found by quantifying the Respiration-to-Noise Ratio (RNR): 

\begin{equation}
\begin{aligned}
    RNR(\tau) &= \frac{  \int_{0.1}^{0.6} { \lvert R(f,\tau) \rvert }^{2} d f }{\int_{0}^{0.1}  { \lvert R(f,\tau) \rvert }^{2} d f + \int_{0.6}^{f_s/2} { \lvert R(f,\tau) \rvert }^{2} d f}
\end{aligned}
\end{equation}

where $R(f,\tau)$ is the Fourier transform of the receiving signal across slow time. The range bin exhibiting the peak RNR is designated as the position of the body. The observation range of interest, delineated as $[\tau_{0} - \tau_{max}, \tau_{0} + \tau_{max}]$, is obtained by incorporating a brief propagation offset, accounting for adjacent anatomical structures.

\subsubsection{Activity signal generation} \label{motion}

\begin{figure}
\centering
\includegraphics[width=1.05\linewidth]{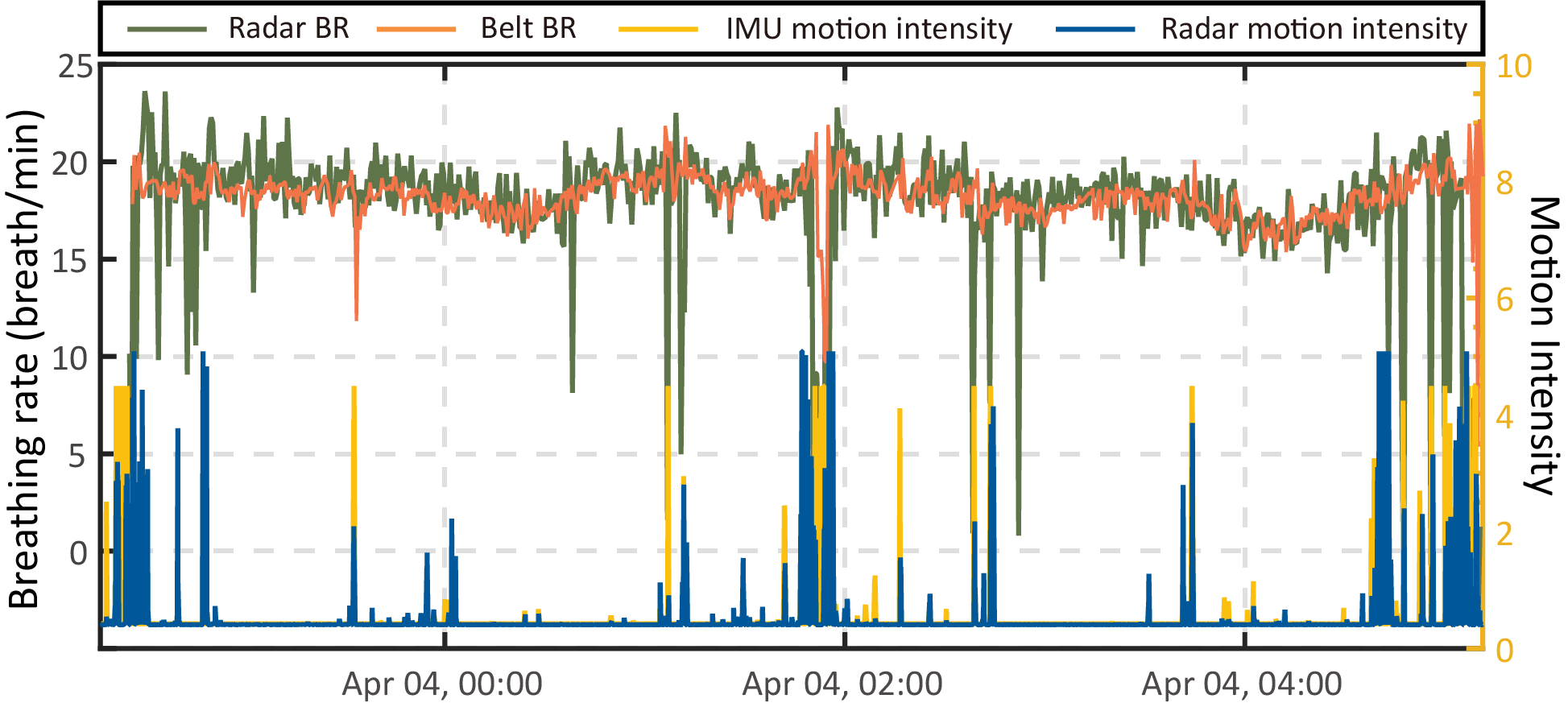}
\caption[Overnight respiration comparison.]{Overnight respiration rate comparison between radar and PSG belt channel, and comparison of overnight activity signals as detected by radar and IMU. }
\label{fig:radar_vs_PSG_breath}
\vspace{-2mm}
\end{figure}

The magnitude of the reflected radar energy is strongly modulated by large body movement, whereas displacements due to cardiorespiratory motion are much smaller. We therefore use the temporal variance of the received energy as a motion marker. For each slow-time index $t$, we compute the total received energy within the range of interest:
\begin{equation}
E(t)=\int\left|r(t, \tau)\right|^2 \mathrm{d}\tau,
\end{equation}
and estimate its local variance $\operatorname{Var}_L\{E(t)\}$ using a sliding window of length $L{=}1$~s. Motion is detected when $\operatorname{Var}_L\{E(t)\}$ exceeds a robust threshold defined as
\[
\text{median}(E) + \lambda\,\text{MAD}(E),\quad \lambda = 6,
\]
where $\text{MAD}$ denotes the median absolute deviation, We set $\lambda = 6$, which corresponds to approximately four standard deviations in a Gaussian distribution. As illustrated in Fig.~\ref{pre-processing}C, epochs containing motion artefacts exhibit pronounced spikes in variance compared with stable periods and can thus be reliably identified.

\subsubsection{Respiratory signal reconstruction} 

Respiration-related motion is encoded primarily in the phase of the clutter-suppressed radar signal within the observation range identified in Section~\ref{Position}. Let $\tilde{r}(t,\tau)$ denote the complex-valued, clutter-removed radar signal at slow time $t$ and fast time (range) $\tau$. For each range bin within $[\tau_{0} - \tau_{\max}, \tau_{0} + \tau_{\max}]$, we compute the instantaneous phase and unwrap along the slow-time dimension to obtain a continuous phase trajectory:
\begin{equation}
    \tilde{\phi}(t,\tau) = \text{unwrap}\big(\arg\{\tilde{r}(t,\tau)\}\big),
\end{equation}
where $\arg(\cdot)$ denotes the four-quadrant phase operator. $\tilde{\phi}$ is proportional to the displacement of the chest wall along the radar line-of-sight.

The human torso occupies several adjacent range bins, and each range bin within the observation window contains a slightly different, partially modulated version of the underlying respiratory motion. As illustrated in Fig.~\ref{pre-processing}D, phase demodulation of each range bin yields a family of one-dimensional displacement signals $\tilde{\phi}(t,\tau_k)$, $k = 1,\dots,K$, all of which contain respiration together with noise and multipath. 

To this end, we employ multivariate empirical mode decomposition (MEMD), an extension of empirical mode decomposition to multichannel data. We exploit the redundancy across range bins and jointly extract the dominant respiratory mode. Let 
\[
    \mathbf{x}(t) = \big[\tilde{\phi}(t,\tau_1), \tilde{\phi}(t,\tau_2), \dots, \tilde{\phi}(t,\tau_K)\big]^\top
\]
denote the $K$-dimensional vector comprising all unwrapped phase signals within the participant’s range window. MEMD adaptively decomposes $\mathbf{x}(t)$ into a set of multivariate intrinsic mode functions (IMFs),
\begin{equation}
    \mathbf{x}(t) = \sum \mathbf{c}_m(t) + \mathbf{r}(t),
\end{equation}
where each $\mathbf{c}_m(t) \in \mathbb{R}^K$ is an IMF with well-defined instantaneous frequency shared across channels, and $\mathbf{r}(t)$ is a slowly varying residual. We identify the respiratory component by selecting the IMF whose instantaneous frequency lies within the physiological breathing band (6–36 breaths/min) and maximizes the average Respiration-to-Noise Ratio across range bins. The selected IMF, $\mathbf{c}_{\text{resp}}(t)$, is then collapsed into a single surrogate respiratory waveform by averaging over channels:
\begin{equation}
    R(t) = \frac{1}{K} \sum_{k=1}^{K} c_{\text{resp},k}(t),
\end{equation}
and finally band-pass filtered between 0.1 and 0.6~Hz to further suppress residual noise and drift.

Fig.~\ref{pre-processing}E and Fig.~\ref{fig:radar_vs_PSG_breath} illustrate representative examples of the reconstructed radar respiratory signal $R(t)$ compared with the thoracic belt signal. The two signals exhibit highly similar morphology and accurate breathing rate tracking on overnight timescales. It is consistent with the fact that both sensors capture the same underlying chest-wall motion.

\subsection{Neural Network Design}

\begin{figure*}
\centering
\includegraphics[width=1\textwidth]{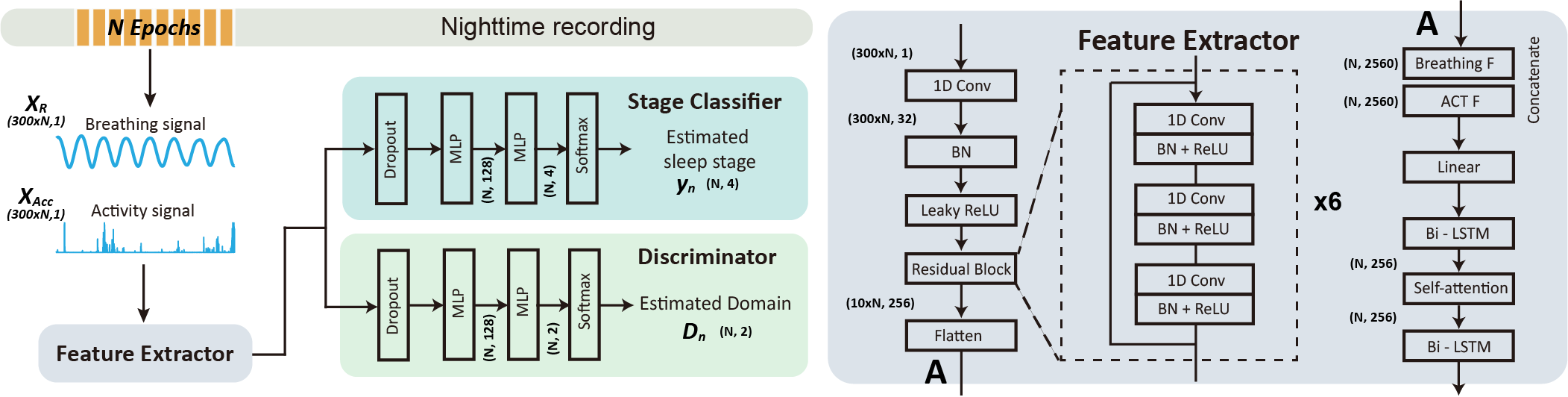}
\caption{ Model overview. The end-to-end network comprises a feature extractor $\mathcal{F}$, a sleep-stage classifier $\mathcal{C}$, and a domain discriminator $\mathcal{D}$. }
\label{Flowchart_machinelearning}
\end{figure*}

Fig. \ref{Flowchart_machinelearning} illustrates the neural network structure of the proposed end2end Sleep Stage Classification network, which includes a feature extractor $\mathcal{F}$, a discriminator $\mathcal{D}$, and a stage classifier $\mathcal{C}$. The feature extractor is responsible for identifying sleep-related features from the input signals, while the stage classifier categorizes these features into the corresponding sleep stages (W, REM, LS, and DS). The discriminator distinguishes between input data originating from either radar or PSG signals.

\paragraph{Inputs and task}
We use two radar-derived channels---respiration and actigraphy (10\,Hz)---and segment the night into $L_{\text{seq}}$-epoch sequences (30\,s per epoch). The goal is to predict four sleep stages (W/REM/LS/DS) per epoch. To improve generalisation, the network is trained jointly on a large PSG source domain (belt respiration + wrist IMU) and a small radar target domain.

\paragraph{Feature extractor \(\mathcal{F}\)}
The design of the feature extractor employs a combined CNN-RNN architecture. The CNN is used to extract features from each epoch's respiratory and motion signals, while the RNN captures temporal relationships across epochs to model dynamic features. The design of the CNN architecture is inspired by residual neural networks (ResNet), which comprises six stacked bottleneck residual blocks. 
Each block adopts a pre-activation bottleneck structure with $1{\times}1$ to $3{\times}1 (dilated)$ to $1{\times}1$ convolutions, batch normalisation and leaky rectified linear units (Leaky ReLU). Strides and dilation factors increase across blocks ($s{=}1,2,2,2,2,2$ and $d{=}1,1,2,4,8,16$), expanding the receptive field while progressively reducing temporal resolution. Channel widths grow from 64 to 256, enabling a rich representation with modest computational cost. Residual connections facilitate gradient flow and enhance feature propagation.

The CNN outputs for respiration and actigraphy are concatenated channel-wise and linearly projected to a 256-dimensional feature vector per epoch. To capture temporal dependencies across epochs, these per-epoch features are then passed through a two-layer bidirectional long short-term memory (Bi-LSTM) network. A time-wise attention mechanism is inserted between the Bi-LSTM layers to assign adaptive weights to different epochs, emphasising the most informative temporal contexts for stage classification. Dropout layers are applied throughout the CNN and Bi-LSTM components to reduce overfitting.

\paragraph{Heads $\mathcal{C}$ and $\mathcal{D}$.}
The stage classifier $\mathcal{C}$ is a two-layer multilayer perceptron (MLP) with a final softmax layer, producing class posteriors
\begin{equation}
p_\theta(y\mid x)=\mathrm{softmax}\bigl(\mathcal{C}(\mathcal{F}(x))\bigr),
\end{equation}
where $y\in\{1,\dots,4\}$ indexes sleep stages. The domain discriminator $\mathcal{D}$ is a two-layer MLP with a sigmoid output
\begin{equation}
d(x)=\sigma\bigl(\mathcal{D}(\mathcal{F}(x))\bigr)\in(0,1),
\end{equation}
where label 1 denotes target (radar) and 0 denotes source (PSG).

\subsection{Domain Adaptation in Sleep Staging}

\begin{figure}
\centering
\includegraphics[width=0.49\textwidth]{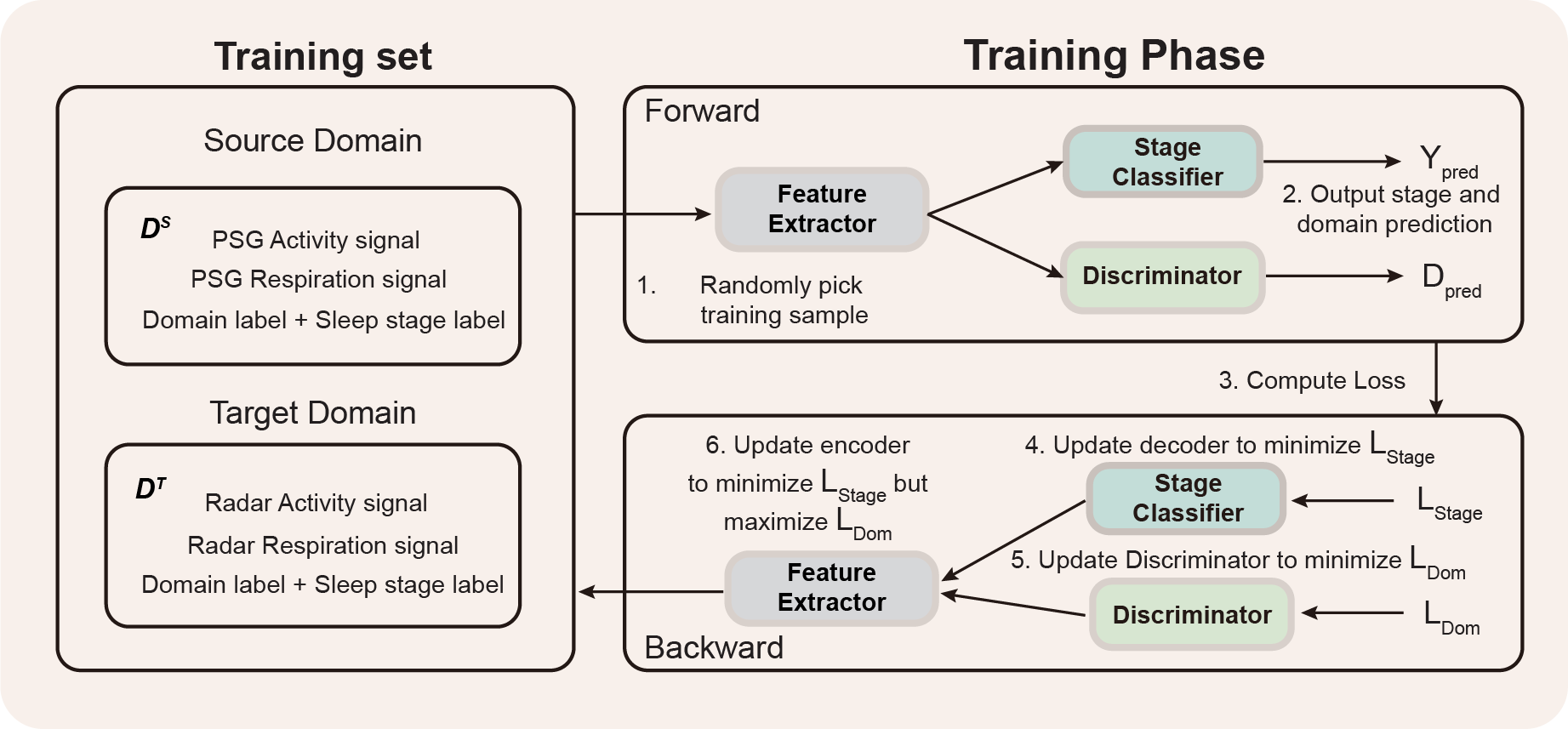}
\caption{ The training framework using the proposed domain adaptation. In the training, a feature extractor is trained to produce domain-invariant features by transferring knowledge from source to target. The network achieves the goal via a minimax game between the feature extractor and discriminators without labeled samples in the target domain.}
\label{Flowchart_training}
\end{figure}

Fig. \ref{Flowchart_training} outlines the proposed training framework. We attach a domain-adversarial module to the shared encoder $\mathcal{F}$ to align feature distributions between the source (PSG-derived respiration/activity) and target (radar-derived signals) while keeping features discriminative for sleep staging. Let $D^s{=}\{(x_s,y_s)\}$ and $D^t{=}\{(x_t,y_t)\}$ denote labeled PSG and radar samples, respectively, with $y\!\in\!\{1,\dots,N_c\}$ (one-hot encoded). During training, a gradient-reversal layer (GRL) in front of $\mathcal{D}$ multiplies the back-propagated gradient by $-\lambda_{\text{adv}}$, forcing $\mathcal{F}$ to produce domain-invariant representations that confuse $\mathcal{D}$ but remain useful for the stage classifier $\mathcal{C}$.

\paragraph{Loss calculation}
We minimize the class-weighted cross-entropy on both domain to mitigate label imbalance:

\begin{equation}
\begin{aligned}
\mathcal{L}_{\mathrm{cls}} = &-\mathbb{E}_{(x,y)\sim P}\sum_{c=1}^{N_c} w_c\, y^c \log p^c,\quad 
\end{aligned}
\end{equation}

where $p=\mathcal{C}(\mathcal{F}(x))$ are class posteriors and $\{w_c\}$ are class weights to mitigate label imbalance.
Class weights are calculated to be inversely proportional to class frequencies, to mitigate label imbalance in four-stage scoring.

To reduce residual distributional shifts between PSG and radar features, we train $\mathcal{D}$ to discriminate the domain
of features, while $\mathcal{F}$ is trained to confuse $\mathcal{D}$ via a gradient reversal layer (GRL). The discriminator binary cross-entropy loss is:

\begin{equation}
\begin{aligned}
\mathcal{L}_{\mathrm{dom}} ={}\;
&-\mathbb{E}_{x_s\sim P_s}
\log\!\left(1-\mathcal{D}(\mathcal{F}(x_s))\right) \\
&-\mathbb{E}_{x_t\sim P_t}
\log\!\left(\mathcal{D}(\mathcal{F}(x_t))\right).
\end{aligned}
\end{equation}

We implement adversarial alignment using a gradient reversal layer (GRL): $\mathcal{D}$ is optimized to \emph{minimize} $\mathcal{L}_{\mathrm{dom}}$, whereas $\mathcal{F}$ receives the negated gradient, effectively \emph{maximizing} $\mathcal{L}_{\mathrm{dom}}$.

\paragraph{Label-aware alignment}
To avoid class mixing under pure domain confusion, we add a class-conditional alignment. We adopt a label-aware contrastive term that pulls together features from the same class across domains and pushes apart different classes:

\begin{equation}
\begin{aligned}
\mathcal{L}_{\mathrm{align}} ={}\;
\mathbb{E}_{i,j}\Big[
&\mathbf{1}_{y_i=y_j}\,\|z_i-z_j\|_2^2 \\
&+ \mathbf{1}_{y_i\neq y_j}\,
\max\!\left(0,\, m-\|z_i-z_j\|_2\right)^2
\Big].
\end{aligned}
\end{equation}

with $z{=}\mathcal{F}(x)$ and margin $m{>}0$. Pairs are sampled across $(D^s,D^t)$ to encourage cross-domain class compactness.

\paragraph{Overall objective}
We solve

\begin{equation}
\begin{aligned}
\hat{\mathcal{D}} &= \min_{\mathcal{D}}\; \mathcal{L}_{\mathrm{dom}}, \\
\hat{\mathcal{F}},\hat{\mathcal{C}} &= 
\min_{\mathcal{F},\mathcal{C}}\big(
\mathcal{L}_{\mathrm{cls}}
-\lambda_{\mathrm{adv}}\mathcal{L}_{\mathrm{dom}}
+\lambda_{\mathrm{align}}\mathcal{L}_{\mathrm{align}}
\big).
\end{aligned}
\end{equation}

Following common practice in adversarial adaptation, $\lambda_{\mathrm{adv}}$ is ramped up with training progress $p\!\in\![0,1]$ as
$\lambda_{\mathrm{adv}}(p)=\tfrac{2}{1+\exp(-\gamma p)}-1$ with $\gamma{=}10$, balancing early discriminative learning and later alignment. We set $\lambda_{\mathrm{align}}$ to a small constant $0.1$.

\paragraph{Training procedure.}
At each iteration:  
(1)~a mixed mini-batch of PSG and radar samples is drawn;  
(2)~the feature extractor produces per-epoch embeddings that are forwarded to the stage classifier and discriminator;  
(3)~the discriminator parameters are updated to minimize $\mathcal{L}_{dom}$;  
(4)~the encoder and classifier parameters are updated to minimize $\mathcal{L}_{\mathrm{cls}}-\lambda_{\mathrm{adv}}\mathcal{L}_{\mathrm{dom}}+\lambda_{\mathrm{align}}\mathcal{L}_{\mathrm{align}}$ through the GRL.  
  
This adversarial co-training enables the encoder to preserve sleep-stage discriminability while making the learned feature distribution invariant to sensing modality differences.

\subsection{Decisions Ensemble and Confidence level}
Since the seq2seq network is a multiple-output network, advancing the input sequence of size $L_{\text{seq}}$ by one epoch when evaluating it on a test recording will result in an ensemble of $L_{\text{seq}}$ decisions at every epoch. Fusing this decision ensemble leads to a final decision which is usually better than individual ones. We use the multiplicative aggregation scheme to ensemble all decisions \cite{phan2019seqsleepnet}:

\begin{equation}
\log P\left(y_m\right)=\frac{1}{L_{\text{seq}}} \sum_{i=m-L_{\text{seq}}+1}^m \log P \left(y_m \mid {f}_{m}\right)
\end{equation}

Eventually, the predicted label $\hat{y}_m$ is determined by likelihood maximization:

\begin{equation}
\hat{y}_m=\underset{y_m}{\operatorname{argmax}} \log P\left(y_m\right)
\end{equation}

The cross-entropy of the probability distribution is used to measure the network's intrinsic uncertainty \cite{cover1999elements}. The confidence  $\gamma$ of the network can be represented as \cite{phan2022sleeptransformer}:

\begin{equation}
\begin{aligned}
H_{\text {normalized }}(\hat{y}) &=-\frac{1}{\log (4)} \sum_{c=1}^4 \hat{y}_c \log \left(\hat{y}_c\right), \\
 \gamma(\hat{{y}})&=1-H(\hat{{y}})
\end{aligned}
\end{equation}
where $\hat{y}$ is the predicted probability for class $c$. The normalization factor $-\frac{1}{\log (4)}$ ensures that the normalized entropy value lies between 0, indicating full confidence, and 1, indicating complete uncertainty.

\subsection{Model training and Evaluation}

Given the limited sample size of the radar dataset ($N$ = 47), we employed a 5-fold cross-validation approach to comprehensively assess the performance of the proposed algorithm while avoiding potential test set bias. In each iteration, 80\% of the dataset was used for training, with the remaining 20\% reserved for testing. This approach ensures that the model's performance is robustly evaluated across multiple data splits.

To further validate the effectiveness of transfer learning, we implemented various training strategies:
\begin{itemize}
    \item Training Strategy 1: Trained and tested on Surrey cohorts.
    \item Training Strategy 2: Trained on STAGES and Surrey cohort without domain adaptation, and tested on Surrey cohort.
    \item Training Strategy 3: Trained on STAGES and Surrey cohorts with domain adaptation, and tested on Surrey cohorts. 
    \item Training Strategy 4: Trained on STAGES and bedside radar position in the Surrey cohorts with domain adaptation, tested on overhead radar position in the Surrey cohort. 
\end{itemize}

We also explored the impact of using different input modalities by training the network separately on respiratory signals, movement signals, or a combination of both. Additionally, we investigated how varying the input sequence length affected the model's performance. Five segment durations were tested: 120 seconds, 240 seconds, 480 seconds, 960 seconds, and 1920 seconds, corresponding to 4, 8, 16, 32, and 64 sleep epochs, respectively.

Given the class imbalance inherent in sleep stage classification, optimizing solely for overall accuracy would bias results toward the most common sleep stages. To address this, we evaluated the model's performance using several metrics, including F1 scores, overall accuracy, and Cohen's Kappa ($k$). The F1 score was calculated for each class individually, while accuracy and Cohen's Kappa were computed across all classes.

Apart from sleep stage classification, we compared several sleep parameters between the model's predictions and PSG annotations for each participant. These parameters included total sleep time (TST), wakefulness after sleep onset (WASO), sleep onset latency (SOL), sleep efficiency (SE), REM sleep latency, and the distribution of the four sleep stages. These metrics are commonly used indicators of sleep quality and sleep architecture.

The model was trained using the Adam optimizer with an initial learning rate of 0.001, which was halved every 10 epochs. The training was terminated if the model's performance showed no improvement over 15 consecutive epochs.
The training was performed using an RTX 4090 24GB GPU.
\section{Results}

\begin{table*}
\centering
\caption{Sleep stage Classification Accuracy Under Different Training Strategies}
\begin{threeparttable}
\begin{tabular}{ccccccccc}
\toprule

\multicolumn{2}{c}{Training Strategy} & \multirow{2}{*}{ \makecell[c]{Adversarial \\module }} & \multicolumn{4}{c}{F1 Score  (Mean $\pm$ Std) } & \multirow{2}{*}{Accuracy (\%)} & \multirow{2}{*}{Kappa $k$}\\ 

\cmidrule{1-2} \cmidrule{4-7}

Training set & Test set &  &  W  & REM  & LS & DS &  &\\
  
\midrule 

All Radar & All Radar & Disabled & 0.79 $\pm$ 0.10 & 0.65 $\pm$ 0.21 & 0.75 $\pm$ 0.07 & 0.71 $\pm$ 0.27 & 74.99 $\pm$ 6.5 & 0.59 $\pm$ 0.10\\

STAGES + Radar & All Radar & Disabled & 0.73 $\pm$ 0.12 & 0.63 $\pm$ 0.23 & 0.67 $\pm$ 0.09 & 0.68 $\pm$ 0.26 & 71.05 $\pm$ 7.3 & 0.56 $\pm$ 0.13 \\ 

STAGES + Radar & All Radar & Enabled & $\mathbf{0.84 \pm 0.09}$ & $\mathbf{0.71 \pm 0.17}$ & $\mathbf{0.78 \pm 0.07}$ & $\mathbf{0.72 \pm 0.15}$ & $\mathbf{79.5 \pm 6.6}$ &  $\mathbf{0.65 \pm 0.09}$ \\

\makecell[c]{ STAGES + \\ Bedside Radar}&  \makecell[c]{ Overhead \\ Radar} & Enabled & 0.81 $\pm$ 0.09 & 0.69 $\pm$ 0.20 & 0.75 $\pm$ 0.07 & 0.72 $\pm$ 0.26 & 76.84 $\pm$ 6.5 & 0.61 $\pm$ 0.09\\

\hline
\end{tabular}
\label{Training Strategies}
\end{threeparttable}
\end{table*}

\begin{figure*}

\centering
\includegraphics[width=0.99\textwidth]{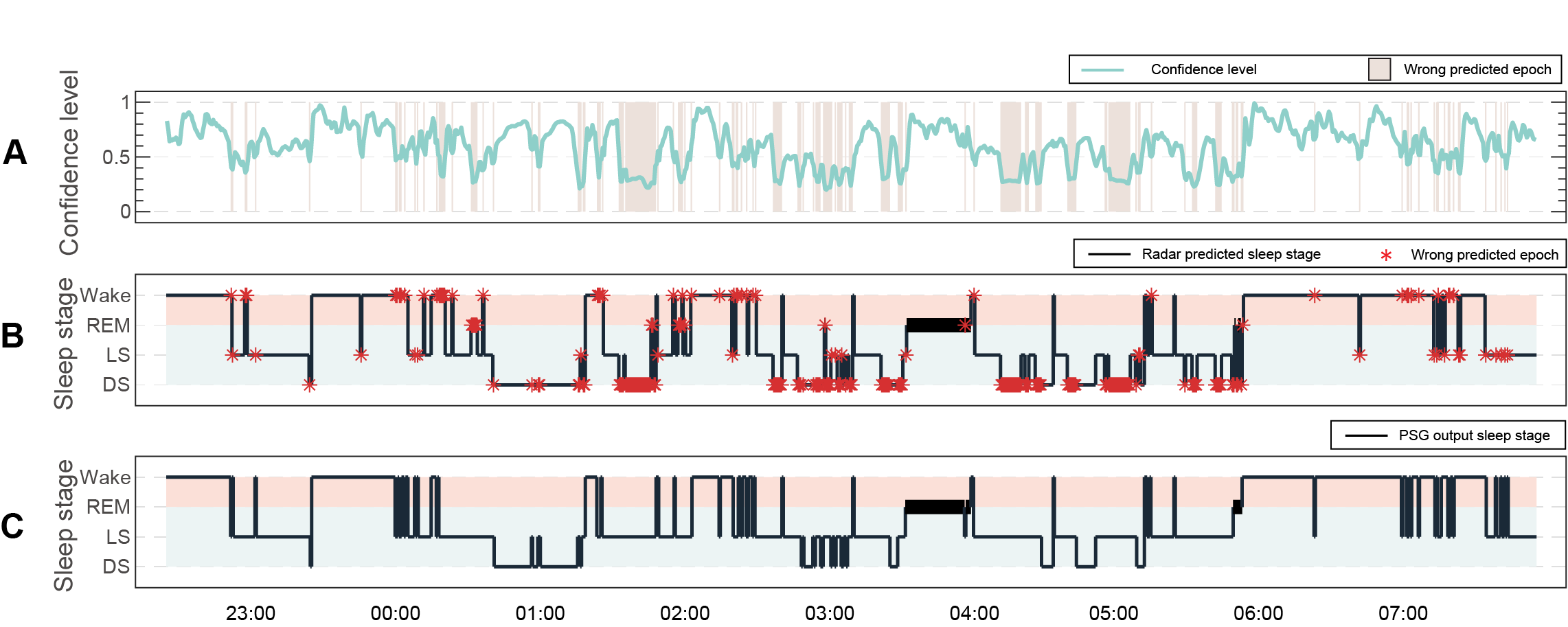}
\caption{ 
Sleep stage classification performance for one participant (average accuracy: 79.57\%). (A) The confidence level of the radar sleep stage classification, with areas shaded representing the misclassification. (B and C) Sample hypnograms for the sleep stage classification from radar predictions and PSG, respectively. These hypnograms cover the analysis period from lights-off to lights-on, with red shading indicating wakefulness and blue shading representing sleep. REM sleep epochs are delineated with thick black lines. }
\label{Radar hypnogram}

\end{figure*}

\subsection{Sleep stage classification accuracy}

Table \ref{Training Strategies} presents the W, REM, LS, and DS classification accuracies achieved under different data combinations and training strategies. The input sequence length was set to 32 epochs.

The results revealed that integrating additional respiratory and IMU data from the large-scale sleep study significantly enhanced sleep stage classification accuracy on the small-scale radar dataset. The sleep stage classification accuracy increased by approximately 5\% compared to models trained solely on radar data.
Moreover, the incorporation of domain adaptation played a crucial role in boosting accuracy. Given that radar data accounted for only 6\% of the entire training set, the model's accuracy declined without the domain adaptation, underscoring the importance of aligning the domains between PSG and radar data. The adversarial learning module effectively facilitated knowledge transfer from the PSG dataset to the radar dataset, thereby boosting model accuracy.

Furthermore, the comparable accuracy across different radar positions suggests that the network did not overfit the small-scale radar dataset. Instead, it learned sleep-related features that were invariant to the specific measurement environment, which indicates robustness in the model's ability to generalize across different setups. 

Given that Training Strategy 3 resulted in the highest overall accuracy, it was selected as the primary training strategy for further analysis and result presentation. A representative hypnogram of the model-predicted and expert-annotated sleep stages over the course of the night is shown in Fig. \ref{Radar hypnogram}.

\subsection{Role of the Proposed Adversarial Discriminator}

\begin{figure}
\centering
\includegraphics[width=0.5\textwidth]{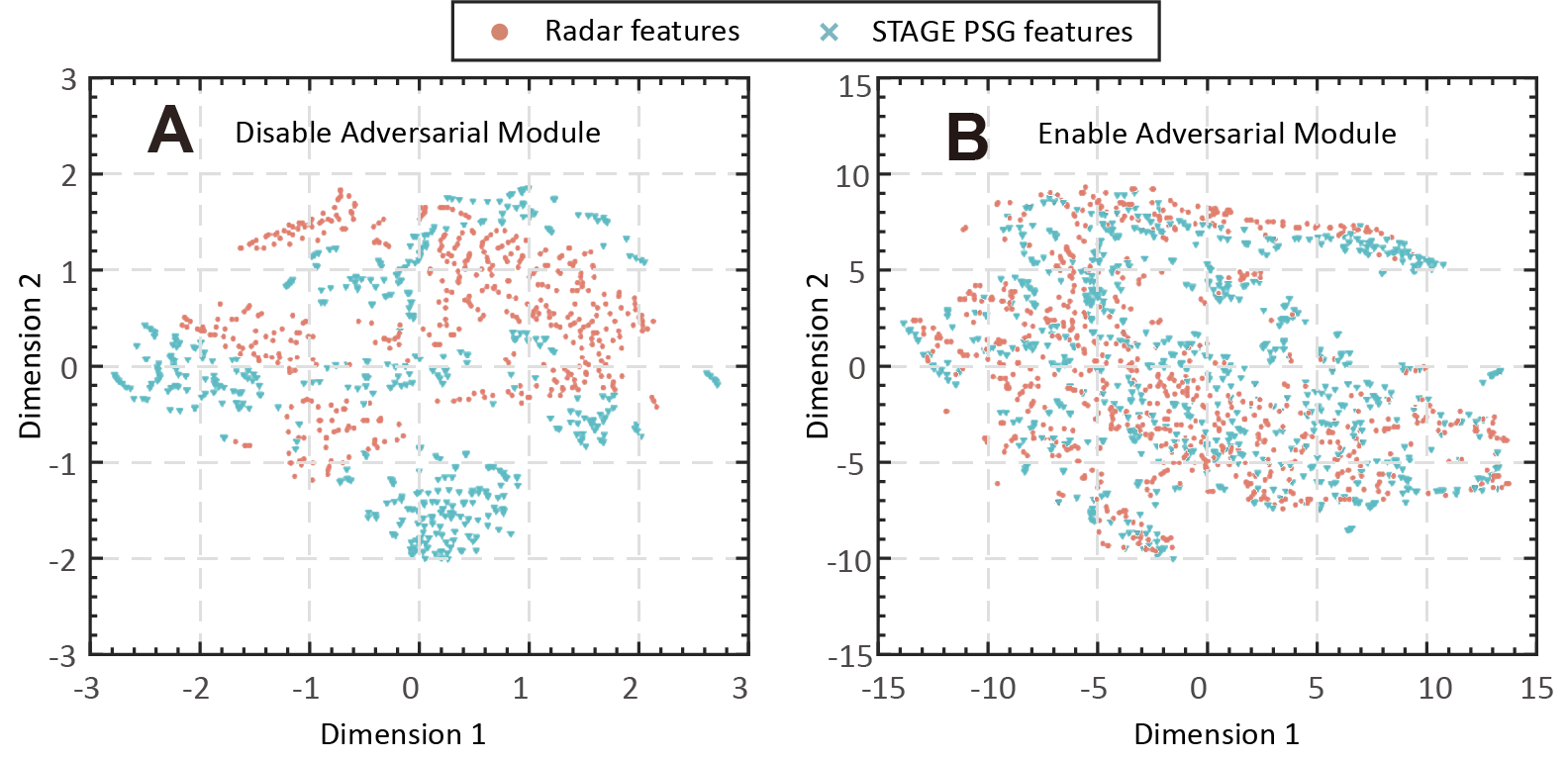}
\caption{ Radar and PSG domain t-SNE visualization of feature vector output from Feature extractor \(\mathcal{F}\). A: Adversarial Module Disabled. B: Adversarial Module Enabled. }
\label{tsne1}
\end{figure}

The role of the proposed adversarial discriminator in learning transferable features for predicting sleep stages was evaluated. Training strategy 2 (non-adversarial) and 3 (with-adversarial) are used for comparison. The t-distributed neighbor embedding (t-SNE) method was employed to visualize the outputs from the feature extractor module. By color-coding the sources, Fig. \ref{tsne1} illustrates that our learned features have almost the same distribution across different sources, while the baseline model learns separable features.

\begin{figure}
\centering
\includegraphics[width=0.5\textwidth]{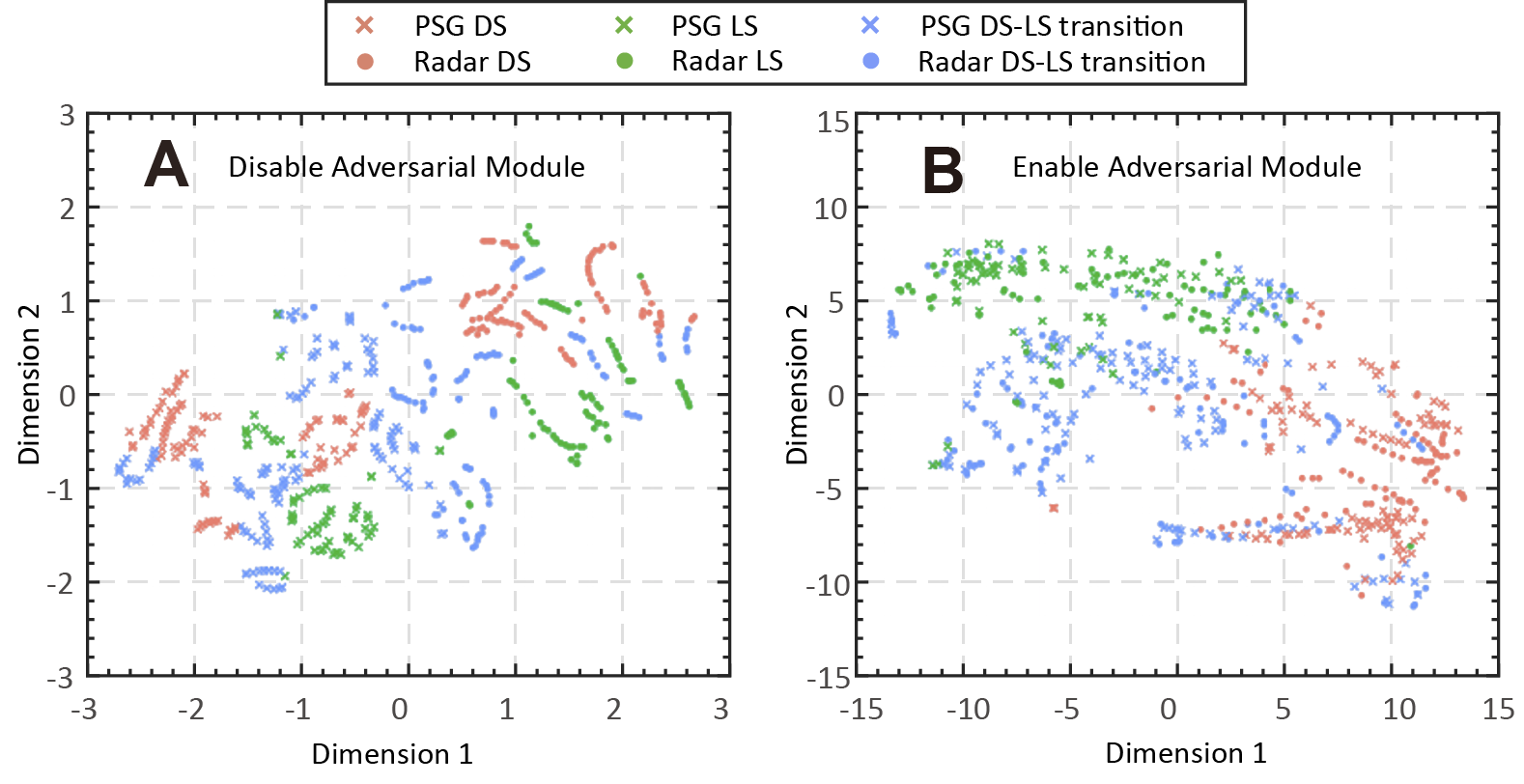}
\caption{t-SNE visualization of learned posterior distribution from Radar and PSG domain. A: Adversarial Module Disabled. B: Adversarial Module Enabled. }
\label{tsne2}
\end{figure}

Next, we illustrate the benefits of conditioning on the posterior distribution and its ability to recover underlying concepts not specified in the labels. We consider the learned features for transition periods between light and deep sleep, which might be a class different from both light and deep sleep. We define transition periods as epochs with both light and deep sleep as neighbours and visualize them with a different colour. Color-coding stages and shape-coding sources, Fig. \ref{tsne2} shows that the learned features from transition periods are segregated, similar to those from light and deep sleep. This indicates that our learned features have recovered the concept of a transition period, which is helpful in understanding and predicting sleep stages.

Transition‑rich nights may reflect underlying sleep instability. This characteristic could potentially serve as a prospective biomarker for assessing sleep stability or related neurophysiological states in future research.

\subsection{Impact of different input signals}

\begin{figure}
\centering
\includegraphics[width=0.5\textwidth]{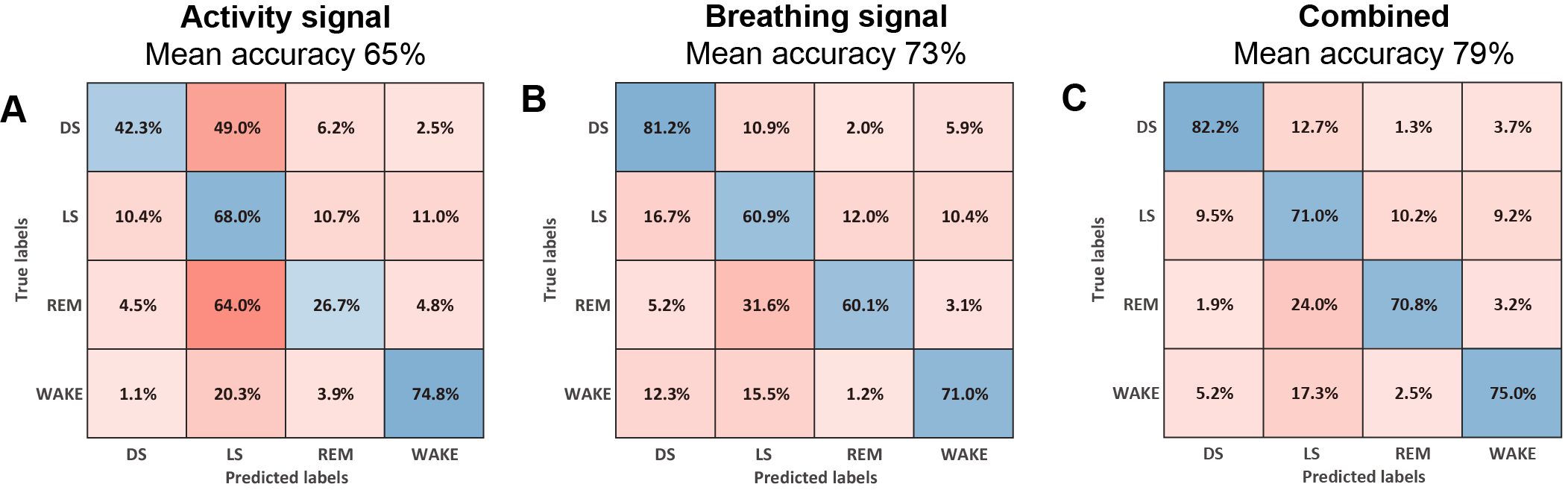}
\caption{ 
Sleep stage classification performance for different input signals: (A) using the activity signal as input, (B) using the breathing signal as input, and (C) combining both signals as input. }
\label{Radar input acc}
\end{figure}

The proposed model is capable of performing sleep stage classification using multiple input channels. Fig. \ref{Radar input acc} displays the results for different network training scenarios: 1) using only motion signals, 2) using only respiratory signals, and 3) combining both types of signals.

The results indicate that models trained with motion signals alone exhibited few false positives in the wake stage but struggled to differentiate other sleep stages, particularly REM sleep and deep sleep. In contrast, incorporating both respiratory and motion signals proved to be the most effective approach, combining the high precision of respiratory signals with the superior wake classification accuracy observed in motion signal-based models. This multimodal input approach led to a 6\% improvement in overall classification accuracy, with root mean square error reductions across nearly all sleep metrics.

\begin{figure}
\centering
\includegraphics[width=0.45\textwidth]{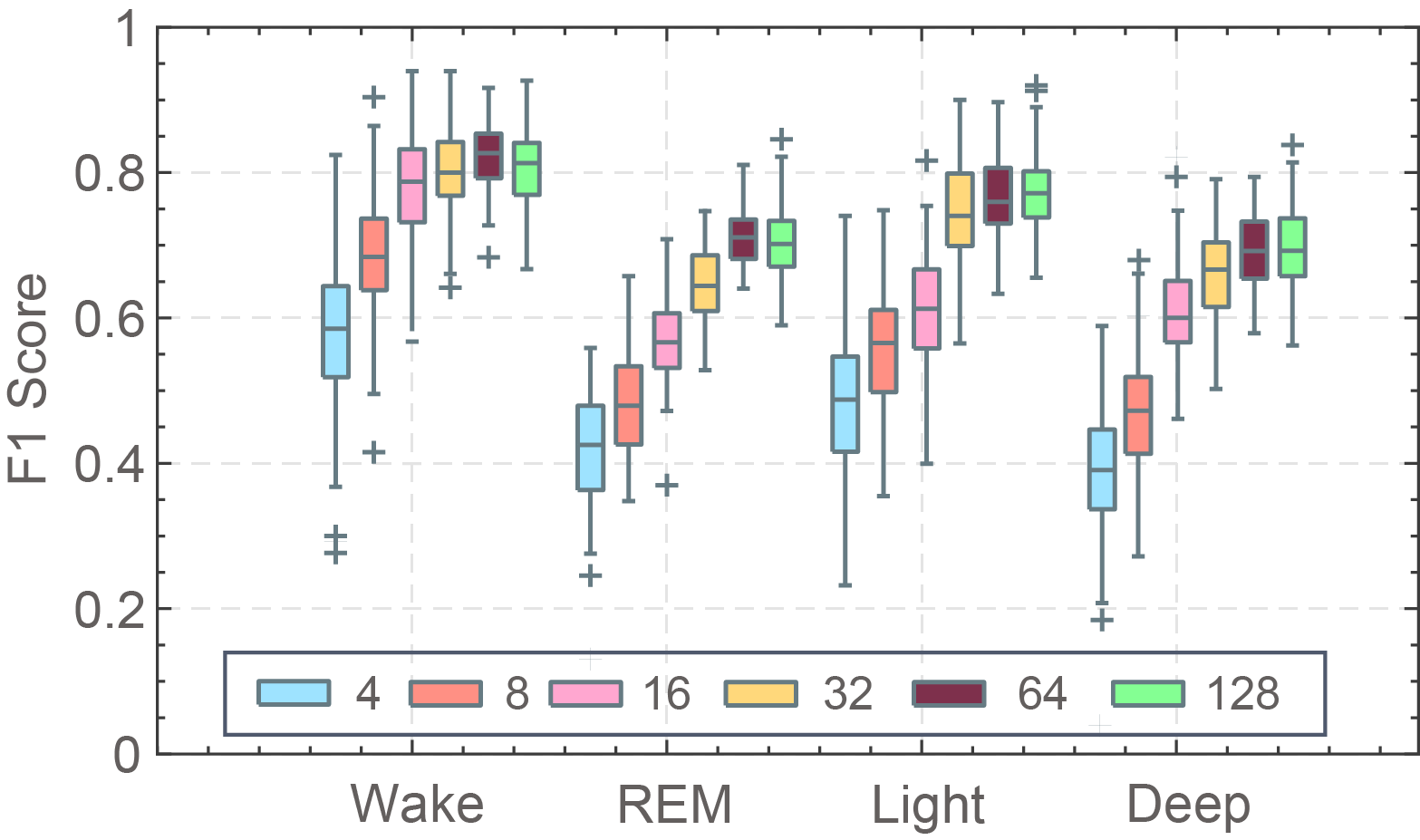}
\caption{ 
Sleep stage classification performance for different input signal lengths. f1 scores for different sleep stages and segment sizes.  }
\label{length input acc}
\end{figure}

To assess the importance of temporal modeling, the deep neural network was trained using different input segment lengths. Fig. \ref{length input acc} illustrates the impact of input sequence length on classification outcomes. The findings suggest that performance improves across all sleep stages with increasing input segment size, with the most pronounced gains observed for smaller window sizes. Once the segment length exceeded 32 epochs, the effect on performance became less significant. It should be noted that a longer input epoch length requires more computational resources and a longer inference latency. Thus, the proper epoch length should be decided according to the purpose of sleep monitoring, considering the tradeoff between performance and computational burden.

\subsection{Sleep parameters accuracy}

\begin{figure}

\centering
\includegraphics[width=0.45\textwidth]{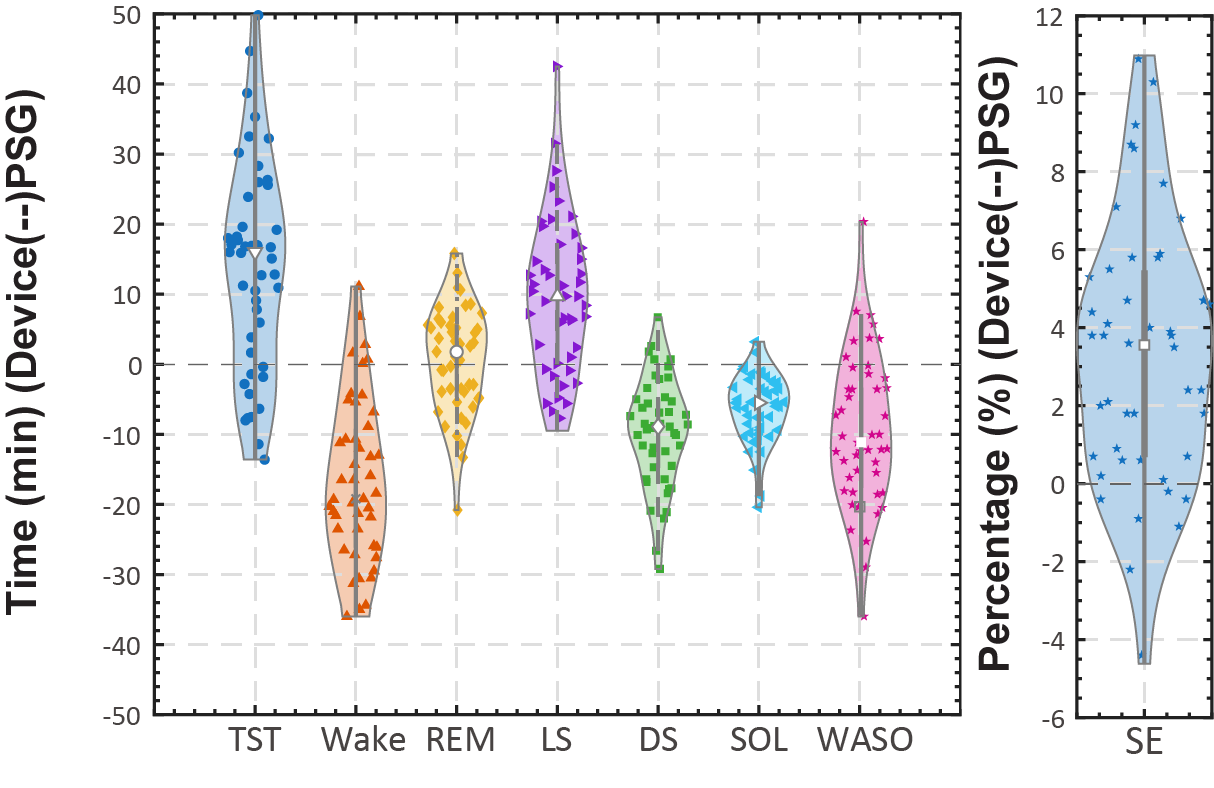}
\caption[Violin plots representing the estimation accuracy for various sleep parameters.]{  Violin plots representing the estimation accuracy for various sleep parameters. The black dotted line indicates the line of no deviation. Data points plotted below this line represent underestimations by radar, while those above the line represent overestimations. Within the central boxplot of each violin, the white marker denotes the median value.}
\label{fig:sleep_param_accuracy2}

\end{figure}

The estimation errors of the radar system after transfer learning are summarized in Figure~\ref{fig:sleep_param_accuracy2}. These metrics were derived from 30-second epoch-level labels and therefore reflect the cumulative effect of staging performance on higher-level sleep indices. Quantitatively, across all 47 nights, the mean ($\pm$SD) radar–PSG difference was 12.3~$\pm$~15.2~min for total sleep time, 9.4~$\pm$~11.1~min for light sleep, $-9.7$~$\pm$~7.7~min for deep sleep, and 3.4~$\pm$~3.4~percentage points for sleep efficiency. Wake after sleep onset and wakefulness were still slightly underestimated ($-13.9$~$\pm$~10.5~min and $-16.9$~$\pm$~11.8~min, respectively), whereas REM sleep and sleep onset latency showed only small residual biases (0.5~$\pm$~7.4~min and $-6.3$~$\pm$~4.8~min).

Despite these biases, the absolute errors remain within clinically acceptable limits. The mean absolute percentage error across all subjects is comparable to or better than previously reported non-contact systems operating in sleep lab environments \cite{lauteslager2020performance,toften2020validation}. Moreover, the radar-derived estimates of sleep efficiency and sleep onset latency showed strong correlation with PSG (Pearson $r = 0.79$ and $r = 0.82$, respectively), suggesting that the radar system reliably captures the overall sleep–wake macrostructure even in the presence of occasional epoch-level misclassifications.

From a clinical standpoint, these deviations are unlikely to affect the interpretation of nightly sleep quality. The radar’s consistent bias direction (slight overestimation of sleep) can be statistically compensated through model calibration or regression correction in future iterations.

\subsection{Effect of health conditions}

\begin{figure*}

\centering
\includegraphics[width=0.95\textwidth]{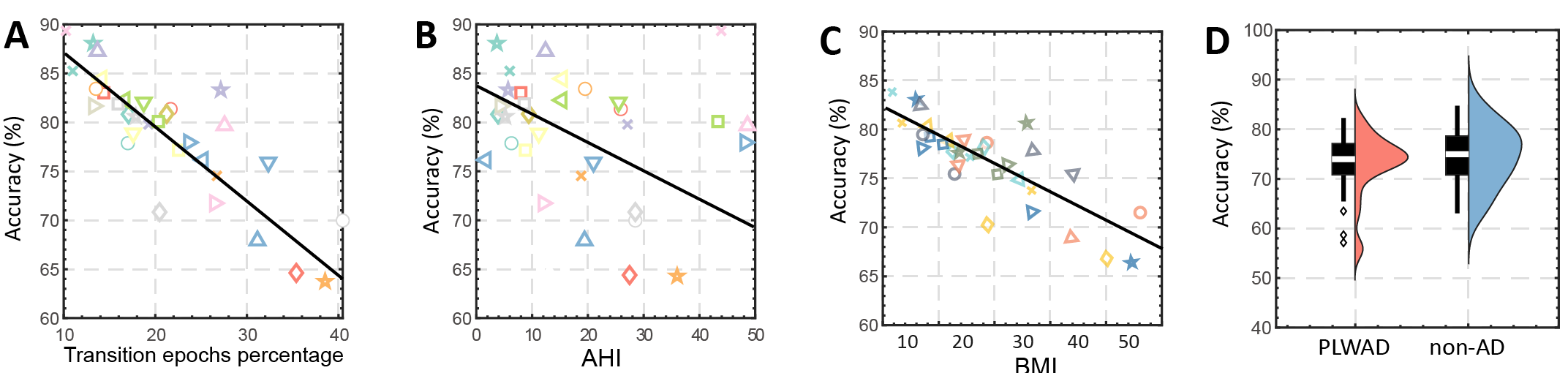}
\caption{ Correlation plot of Radar-Estimated WRLD Accuracy with Various Parameters. A: Accuracy versus transition epochs. B: Accuracy versus AHI. C: Accuracy versus BMI. D: Accuracy for PLWAD and controls. }
\label{bmi_vs_parameters}

\end{figure*}

We employed a linear mixed-effects model to investigate how characteristics of the sleep recording could influence sleep stage classification outcomes, as illustrated in Fig. \ref{bmi_vs_parameters}. One notable observation was the impact of transitioning epochs, which are characterized by sleep stages that differ from their preceding or succeeding neighbors. These epochs are prone to inconsistencies in manual scoring by sleep experts when using polysomnography, and this trend was similarly reflected in the radar-based sleep classification. As the proportion of transitioning epochs increased, the accuracy of WRLD classification decreased, with a 4\% reduction in accuracy observed for every 5\% increase in the proportion of such epochs.

In the 22 sleep records in which the Apnea-Hypopnea index (AHI) was greater than 15 the average WRLD classification was lower than in the sleep records in which the AHI was smaller than 15  (75.2\% vs 83.1\%). An independent samples t-test confirmed that this difference is statistically significant ($p=0.03<0.05$). This suggests that cardiopulmonary irregularities associated with sleep apnea may impair the precision of physiological signals, thereby adversely affecting sleep stage classification.

A negative correlation was observed between Body Mass Index (BMI) and classification accuracy. This trend is likely attributable to the mechanical damping effect of increased subcutaneous adipose tissue, which can restrict the amplitude of chest wall surface displacement during respiration, thereby reducing the signal-to-noise ratio. The potential distortion of the radar-returned respiratory waveform may lead to reduced stage classification accuracy. In contrast, the system demonstrated remarkable stability across AD patients, we found no significant divergence in classification accuracy between PLWAD and healthy controls. This indicates that the proposed transfer learning framework effectively generalizes across these cohorts, maintaining high performance despite the potential behavioral or physiological heterogeneity associated with neurodegenerative conditions.

For radar-based systems to be clinically viable, they must accurately quantify sleep despite the disrupted architectures and frequent transitions typical of aging and neurodegeneration. Although distinguishing sleep stages becomes more difficult in the presence of disorders like apnea, excluding such cases would fail to address the needs of the very populations who stand to benefit most from this technology. Therefore, our validation cohort was inclusive of participants with diverse health profiles and sleep disturbances to ensure the findings reflect real-world performance.

 \section{Discussion}

Our study demonstrates the feasibility of UWB radar-based sleep staging through a novel transfer learning framework, achieving competitive performance (accuracy: 79.5\%, Kappa: 0.65) in a cohort of older adults, including individuals with prodromal or mild Alzheimer’s disease.

Beyond improving cross-domain generalisation, the proposed transfer-learning framework reflects a broader conceptual contribution of this work: it demonstrates that physiological and behavioural dynamics extracted from radar can serve as a unified substrate for sleep analysis across sensing modalities. The framework maps respiration and movement channels to sleep stages, and this mapping is modality-agnostic. In principle, the same architecture can be trained on PSG respiratory belts and actigraphy, radar-derived signals, or any other sensor capturing thoraco-abdominal motion and gross body activity (e.g., under-mattress mats or FMCW/CW radar). Because sleep staging relies primarily on autonomic regulation and behavioural transitions, the physiological principles are shared across modalities, enabling meaningful knowledge transfer. This indicates, first, that the network architecture is effective at extracting modality-invariant representations of sleep, and second, that the features learned from one modality embody stable and transferable physiological knowledge rather than sensor-specific artifacts.

Table \ref{parameter} provides a comparison of recent studies that have utilized radar sensors for sleep stage classification. To the best of our knowledge, our study is the first to focus on radar-based sleep monitoring in an elderly population, including individuals with AD and sleep disorders. It is also one of the few to employ transfer learning using a large-scale PSG dataset to improve radar-based measurements. Although some prior studies have reported higher sleep classification accuracy, these were typically conducted under strictly controlled conditions, where radar placement and orientation were tightly regulated to minimize variability \cite{hong2018noncontact,9403900}. In contrast, our study faced additional challenges due to the diverse age distribution and health conditions of participants. The fragmentation of sleep associated with aging, coupled with irregular respiratory patterns in individuals with sleep apnea, negatively impacted classification accuracy. Our findings reveal that frequent sleep interruptions reduce confidence levels in classification, leading to lower accuracy. Additionally, sleep apnea was found to hinder the extraction of clear respiratory signals, further complicating the classification of sleep stages. Nevertheless, our approach demonstrated accuracy comparable to studies involving younger and healthier cohorts.

\begin{table*}
\centering
\label{compare}
\caption{A Comparison of Performance with Previous Studies}
\begin{threeparttable}
    \begin{tabular}{ccccccc}
\toprule

Study & Radar Processing method & Feature & Classifier & Class & Participant & Performance  \\\midrule

2020 \cite{heglum2021distinguishing} & FFT & MOV, RR & Logistic Regression & W/S &  \makecell[c]{ N = 12, Healthy adult, \\ Age: 23.0 $\pm$ 3.1 }&  \makecell[c]{Accuracy = 92\%, \\ Kappa = 0.81 } \\ \midrule

2021 \cite{toften2020validation} & FFT & MOV, RR & LSTM & W/R/L/D &  \makecell[c]{ N = 71, Healthy adult, \\ Age = 28.9 $\pm$ 9.7} & \makecell[c]{ Accuracy = 76.7\%, \\ Kappa = 0.63} \\\midrule

2021 \cite{de2021radar} & FFT & MOV, RR & SVM, KNN, AdaBoost & W/R/L/D &  \makecell[c]{ N = 32, Children, \\ Age: 2 months to 14 years}&  \makecell[c]{Accuracy = 58.0\%, \\ Kappa = 0.43 } \\ \midrule

2021 \cite{9403900} & Bandpass filter & MOV, RR, HR & Attention-based bi-LSTM & W/R/L/D &  \makecell[c]{ N = 51, Healthy adult, \\ Age = 30 $\pm$ 8.6} & \makecell[c]{ Accuracy = 82.6\%, \\ Kappa = 0.73} \\ \midrule

2024 \cite{10504259} & Spectrogram & End-2-End & CNN+Transformer & W/R/L/D &  \makecell[c]{ N = 290, Healthy adult, \\ Age = 40.5 $\pm$ 12 } & \makecell[c]{ Accuracy = 76\%, \\ Kappa = 0.64} \\ \midrule

Our study & EMD & End-2-End & Seq2seq CNN+LSTM & W/R/L/D &  \makecell[c]{ N = 47, Older people, \\ Age = 71.2 $\pm$ 6.5 \\ PLWAD } & \makecell[c]{ Accuracy = 79.5\%, \\ Kappa = 0.65} \\

\bottomrule

\end{tabular}
\label{parameter}

\begin{tablenotes}
\small
\item Note: AdaBoost, Adaptive Boosting; MOV, Movement; SVM, Support Vector Machine, TCN, Temporal Convolutional Network; W/S, Wake/Sleep; W/R/L/D, Wake/REM Sleep/Light Sleep/Deep Sleep.
\end{tablenotes}
\end{threeparttable}
\end{table*}

Deep learning models typically demand large, consistent datasets to ensure generalization. However, acquiring large-scale radar data with synchronized PSG is resource-intensive, requiring skilled annotation and imposing a significant burden on participants. Consequently, most existing radar studies are restricted to small, healthy cohorts in controlled laboratories, often resulting in models that fail to generalize to new subjects or operational conditions \cite{baumert2023automatic}. Notably, compared to our previous work conducted in a similar clinical environment \cite{yin2025unobtrusive}, the transfer learning framework implemented in this study significantly enhanced classification accuracy. As shown in Table \ref{Training Strategies}, using the domain adaptive technique, training the model using both large-scale respiratory signals from PSGs from open databases and UWB radar data improves the performance, whereas training with UWB radar data alone does not produce satisfactory results. Therefore, the domain adaptation applied in this study not only improves the performance but also reduces the effort of collecting radar data using PSG, thus saving the development time and cost of a radar-based sleep monitoring system. 

While recent works, such as the smartphone-based UWB radar study by Park et al. \cite{10504259}, have explored similar objectives, our approach distinguishes itself in several critical aspects. Both studies employed domain adaptation to leverage PSG-derived respiratory signals, but our adversarial learning framework uniquely aligns feature distributions between PSG and radar domains through a minimax game between the feature extractor and discriminator. In contrast, it utilized contrastive loss to minimize feature discrepancies between PSG spectrograms and radar Doppler maps. While their method emphasizes label consistency, our adversarial approach explicitly reduces domain shift, which is particularly advantageous for small radar datasets (6\% of training data in our study). This distinction is reflected in our model’s ability to generalize across radar placements (bedside vs. overhead) without overfitting, a challenge not explicitly addressed in smartphone-based setups where antenna alignment is constrained.

\subsection{Study Limitations and Future Work}

While the proposed radar-based sleep classification algorithm holds significant promise in accurately quantifying various sleep parameters, there exist a few limitations that should be addressed in future studies. 

First, our analysis was confined to single-night sleep monitoring for elderly individuals and participants with neurodegenerative diseases. This setting does not capture the variability and complexity of sleep patterns of multi-night and long-term monitoring, where longitudinal data would be critical for a comprehensive understanding of sleep behavior.

Additionally, the study focused on nocturnal sleep, neglecting the prevalent daytime napping habits among older populations. The accuracy of our system in monitoring daytime sleep remains to be assessed. Our ultimate long-term aim is to deploy our radar system to provide a comprehensive 24/365 perspective of physiology and sleep within home environments. This would provide new insights as such continuous, longitudinal data is currently not available. This would be of particular interest in assessing sleep disorders and also observing disease progress in neurodevelopmental disorders.

\section{CONCLUSION}

This study evaluated the potential of commercially available UWB radar technology for monitoring sleep patterns in older adults, including individuals with AD and sleep disorders. We developed an end-to-end deep learning model integrating a 1D CNN and LSTM architecture to classify sleep stages and demonstrated that domain adaptation techniques can effectively train this model, even with limited radar data. The results underscore the efficacy of the proposed neural network and domain adaptation approach, showing high accuracy in sleep stage classification. By leveraging domain adaptation, the model was able to transfer knowledge from large-scale PSG data to radar-based sleep monitoring, improving performance while reducing the need for extensive radar data collection. These findings suggest that UWB radar technology, combined with advanced machine learning techniques, offers a promising solution for long-term, non-intrusive home-based sleep monitoring. The successful application of this approach indicates its potential for broader adoption in tracking sleep health and managing sleep disorders among older adults including people living with dementia.

\section{Acknowledgements}

This research was supported by the UK Dementia Research Institute, Care Research and Technology Centre at Imperial College, London, and the University of Surrey, Guildford. We are grateful to the staff of the Surrey Sleep Research Centre for their assistance with data collection. Our appreciation extends to Mr. Giuseppe Atzori and Ms. Marta Messina Pineda for their contributions to PSG sleep stage scoring and running the trial. We would also like to acknowledge the support from SABP and the Surrey Clinical Research Facility. Special thanks to Vikki Revell (SSRC), Ramin Nilforooshan, Jessica True (Surrey and Borders Partnership), Hana Hassanin (Clinical Research Facility) for their essential roles in facilitating study operations and participant care. 

This research has been conducted using the STAGES - Stanford Technology, Analytics and Genomics in Sleep Resource funded by the Klarman Family Foundation. The investigators of the STAGES study contributed to the design and implementation of the STAGES cohort and/or provided data and/or collected biospecimens, but did not necessarily participate in the analysis or writing of this report. The full list of STAGES investigators can be found at the project website. The National Sleep Research Resource was supported by the U.S. National Institutes of Health, National Heart Lung and Blood Institute (R24 HL114473, 75N92019R002).

\bibliographystyle{IEEEtran}
\bibliography{RadarProject}


\end{document}